\documentclass[11pt]{article}

\usepackage{arxiv}

\usepackage[utf8]{inputenc} 
\usepackage[T1]{fontenc}    
\usepackage{hyperref}       
\usepackage{url}            
\usepackage{booktabs}       
\usepackage{amsfonts}       
\usepackage{nicefrac}       
\usepackage{microtype}      
\usepackage{lipsum}		
\usepackage{graphicx}
\usepackage{doi}

\usepackage{amsmath}
\usepackage{array}
\usepackage[square,numbers]{natbib}
\usepackage{caption}
\usepackage{subcaption}
\usepackage{listings}
\usepackage{xcolor}
\usepackage{textcomp}

\lstdefinelanguage{json}{
  basicstyle=\ttfamily\small,
  breaklines=true,
  frame=single,
  backgroundcolor=\color{gray!10},
}

\title{A unified data format for managing diabetes time-series data:
DIAbetes eXchange (DIAX)}

\author{ Elliott C. Pryor \\
         University of Virginia, Center for Diabetes Technology\\
         \texttt{elliott.pryor@virginia.edu} \\
         \And
         Marc D. Breton \\
         University of Virginia, Center for Diabetes Technology\\
         \texttt{mb6nt@virginia.edu} \\
         \And
         Anas El Fathi \\
         University of Virginia, Center for Diabetes Technology\\
         \texttt{fwt9vd@virginia.edu} \\
}

\hypersetup{
pdftitle={A unified data format for managing diabetes time-series data:
DIAbetes eXchange (DIAX)},
pdfauthor={Elliott C. Pryor, Marc D. Breton, Anas El Fathi},
pdfkeywords={First keyword, Second keyword, More},
colorlinks=false,
pdfborder={0 0 0},
}

\begin{document}
\maketitle

\begin{abstract}
Diabetes devices, including Continuous Glucose Monitoring (CGM), Smart Insulin Pens, and Automated Insulin Delivery systems, generate rich time-series data widely used in research and machine learning.
However, inconsistent data formats across sources hinder sharing, integration, and analysis.

We present DIAX (DIAbetes eXchange), a standardized JSON-based format for unifying diabetes time-series data, including CGM, insulin, and meal signals.
DIAX promotes interoperability, reproducibility, and extensibility, particularly for machine learning applications.
An open-source repository provides tools for dataset conversion, cross-format compatibility, visualization, and community contributions.
DIAX is a translational resource, not a data host, ensuring flexibility without imposing data-sharing constraints.

Currently, DIAX is compatible with other standardization efforts and supports major datasets (DCLP3, DCLP5, IOBP2, PEDAP, T1Dexi, Loop), totaling over 10 million patient-hours of data.
\url{https://github.com/Center-for-Diabetes-Technology/DIAX}
\end{abstract}


\section{Introduction}

CGM and connected insulin delivery devices, including Automated Insulin Delivery (AID) systems and Smart Insulin Pens, are now standard in type 1 diabetes (T1D) care, generating rich streams of glucose, insulin, and behavioral data.
These technologies have created new opportunities for data-driven research with rapid growth in the application of machine learning to diabetes, with models being developed for prediction, decision support, and personalization of therapy \citep{klonoff2025,jacobs2024,zhu2021}.
Beyond machine learning, other applications such as digital twin modeling also rely heavily on rich glucose and insulin data \citep{cappon2024twins}.

However, these methods require large volumes of high-quality data from diverse sources.
A major barrier, beyond data availability is format heterogeneity: each dataset typically uses different conventions for time representation, measurement units, and structure, making integration labor-intensive and error-prone.
As a result, many works only consider data from a few sources, limiting generalizability.

Clinical standardization efforts such as the iCoDE-2 project have underscored the challenges of device heterogeneity, inconsistent metrics, and varying levels of data abstraction when integrating diabetes device data \citep{aaron2024}.
DIAX addresses these challenges by providing a standardized, analysis-ready format specifically tailored to glucose--insulin modeling.
Focusing on CGM and insulin, signals central to most machine-learning and modeling tasks, DIAX offers a consistent structure that enables the incorporation of heterogeneous datasets and supports broader population diversity, a key factor in developing robust, generalizable algorithms.
DIAX serves as a lightweight, research-oriented harmonization layer applied after data extraction from clinical or device systems.
As a translational repository, it distributes conversion and analysis tools rather than raw data, preserving flexibility for contributors and enabling straightforward mapping to existing standards when needed.

Several recent efforts have also aimed to consolidate diabetes data, including BabelBetes\footnote{\url{https://github.com/nudgebg/babelbetes}}, DiaData \citep{cinar2025}, MetaboNet \citep{wolff2026}, and the Diabetes Research Hub\footnote{\url{https://drh.diabetestechnology.org/}}.
These tools provide valuable cleaned datasets or conversion pipelines, and BabelBetes in particular addresses similar consolidation goals.
DIAX differs primarily in two respects: it preserves richer per-signal metadata including device type and insulin formulation, which supports reproducibility across heterogeneous datasets; and it bundles a suite of analysis and visualization utilities in Python and MATLAB, enabling users to compute glycemic metrics, generate AGP-style plots, and perform time alignment without additional tooling.
Additionally, unlike some existing tools that assume a fixed sampling frequency, DIAX stores raw time-stamped observations without enforcing temporal alignment, avoiding interpolation artifacts and preserving compatibility with datasets sampled at irregular or varying rates.
Together, these features position DIAX as a research harmonization layer designed to support cross-dataset analysis and ML workflows.

\section{Methods}

To promote interoperability, reproducibility, and extensibility in diabetes research, we developed a standardized JSON-based format for representing diabetes-related time series data.
It accommodates diverse data signals---CGM, insulin, carbohydrate intake, and other physiological or behavioral signals---within a unified structure.

CGM and insulin are the primary fields; insulin is strongly recommended but not required for type 2 or gestational diabetes.
Carbohydrate data is optional, as it is often missing and meal announcements are typically inaccurate \citep{brazeau2013}.
The proposed structure is flexible, allowing new fields such as `glp1' to be added as data sources evolve.

DIAX avoids enforcing a fixed sampling rate or requiring temporal alignment across data types, thereby supporting heterogeneous and incomplete datasets.
Alignment at the data representation level can introduce artifacts or constrain downstream use cases with unnecessary interpolation assumptions.
Instead, DIAX stores raw time-stamped observations and provides flexible utilities for task-specific alignment and interpolation when needed (see Section~\ref{sec:utilities}).

As of this time, several datasets are available to convert to the DIAX format: DCLP3 ($n=125$), DCLP5 ($n=100$), IOBP2 ($n=343$), PEDAP ($n=99$), T1Dexi ($n=404$), and Loop ($n=845$) \citep{breton2020,brown2019,bionicpancreas2022,wadwa2023,riddell2023,lum2021}; all of which are available from jaeb.org, totaling over 10 million patient hours of data.
Conversion utilities from other formats, including BabelBetes, are also provided, enabling researchers to access DIAX's analysis tools without abandoning existing workflows.
More datasets will be available in the future from the authors and community efforts.

\subsection{Data Structure}

DIAX defines a unified JSON-based schema for representing heterogeneous diabetes time-series data.
Each dataset is organized as one JSON file per subject, named according to a consistent convention (e.g., \texttt{subj\_MyTrial\_001-001.json}).

Each DIAX file contains a set of top-level keys corresponding to specific physiological or behavioral signals such as \texttt{cgm}, \texttt{bolus}, \texttt{basal\_rate}, \texttt{basal\_inj}, \texttt{carbs}, \texttt{smbg}, \texttt{heart\_rate}, and others when available.
Each key includes two parallel arrays:
\begin{itemize}
  \item \textit{time} array containing timestamps in ISO 8601 format, either:
  \begin{itemize}
    \item \texttt{Y:m:d H:M:S Z} with timezone information (preferred)
    \item \texttt{Y:m:d H:M:S} without timezone information, in which case it is assumed to be in local timezone
  \end{itemize}
  \item \textit{value} array containing the corresponding measurements.
\end{itemize}

A structured metadata section that provides contextual information about every key, including units, descriptions, and device details.
This metadata ensures transparency in data interpretation and facilitates downstream analysis.
Metadata fields include:
\begin{itemize}
  \item \texttt{unit} -- Definition of what unit the value of the field is
  \item \texttt{description} -- Text description of the data in the field.
    For example ``Continuous glucose monitor readings'' or ``User announced carbohydrate intake''
  \item \texttt{device} -- optional -- Device used to administer or measure the field
  \item \texttt{precision} -- optional -- Level of precision of the measurement
  \item \texttt{insulin} / \texttt{medication} -- optional -- Specific information about the drug.
    For example, insulin aspart or insulin lispro.
\end{itemize}

The only required data stream is \texttt{cgm}, with insulin data (\texttt{bolus}, \texttt{basal\_rate}, \texttt{basal\_inj}) strongly recommended.
Optional fields such as carbohydrates, SMBG measurements, heart rate, or weight can be included when present in the source dataset.
For insulin, \texttt{basal\_rate} is assumed constant until the next entry while \texttt{basal\_inj} is assumed zero between entries, typically reflecting pump and MDI delivery respectively, though \texttt{basal\_inj} is not limited to long-acting doses and can represent any discrete insulin event, hence the importance of the metadata fields.
This extensible structure allows new fields, such as GLP-1 delivery or additional biosensor streams, to be incorporated without modifying the core schema.

\begin{table}[ht]
  \centering
  \caption{Key definitions used in DIAX format.}
  \label{tab:keys}
  \begin{tabular}{lll}
    \toprule
    \textbf{Key} & \textbf{Units} & \textbf{Description} \\
    \midrule
    \texttt{unique\_id}   &                          & Unique identifier for the subject \\
    \texttt{cgm}          & mg/dL                    & CGM values \\
    \texttt{bolus}        & U                        & Insulin boluses (meal or correction) \\
    \texttt{basal\_rate}  & U/h                      & Basal insulin delivery rate \\
                          &                          & \textit{Assumed constant infusion at provided} \\
                          &                          & \textit{rate between samples} \\
    \texttt{basal\_inj}   & U                        & Basal injection (for MDI) \\
                          &                          & \textit{Assumed no basal insulin between samples} \\
    \midrule
    \multicolumn{3}{l}{\textit{Optional Fields}} \\
    \midrule
    \texttt{carbs}        & g                        & Carbohydrate intake \\
    \texttt{carb\_category} & String key based on type & Announced type of meal \\
                          &                          & `HT' -- Hypo treatment \\
                          &                          & `Less' -- less than usual \\
                          &                          & `Typical' -- standard size \\
                          &                          & `More' -- more than usual \\
                          &                          & `Ann' -- simple announcement \\
    \texttt{smbg}         & mg/dL                    & Self-monitored blood glucose measurements \\
    \texttt{hba1c}        & \%                       & Measured HbA1c value \\
    \texttt{heart\_rate}  & bps                      & Recorded heart-rate \\
    \texttt{steps}        & steps per ten seconds    & Recorded steps in a 10 second interval \\
    \texttt{height}       & cm                       & Height of subject \\
    \texttt{weight}       & kg                       & Weight of subject \\
    \bottomrule
  \end{tabular}
\end{table}

\subsection{Implementation}

The proposed structure is implemented in an open-source GitHub repository\footnote{Code available on GitHub: \url{https://github.com/Center-for-Diabetes-Technology/DIAX}}.
While the repository does not host data, it provides tools to convert raw data into a standardized JSON format.
Users must obtain source data, which may have licensing restrictions, before applying conversion scripts.

\begin{figure}[ht]
  \centering
  \includegraphics[width=0.85\textwidth]{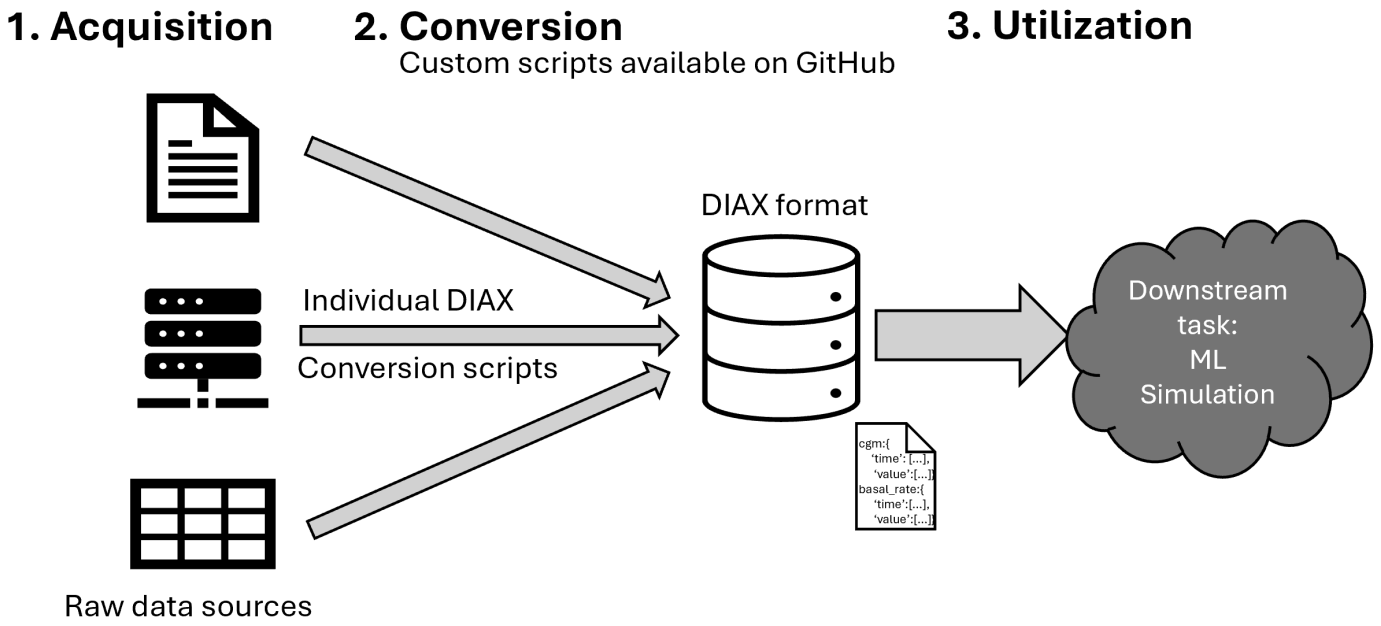}
  \caption{Example conversion workflow.
  Raw data is acquired from the original source.
  Then the matching script from GitHub can be used to convert to DIAX format, which can be used for any downstream task.}
  \label{fig:workflow}
\end{figure}

This model promotes broader dataset inclusion, as sharing a conversion script is often more feasible than uploading entire raw datasets.
By enabling contributions from both researchers and data custodians, the repository fosters a diverse and growing ecosystem of diabetes datasets, enhancing reproducibility and accelerating innovation.
An example workflow is seen in Figure~\ref{fig:workflow}.

\subsection{Utilities}
\label{sec:utilities}

Another main contribution is a set of utilities in Python and MATLAB covering standard use-cases in diabetes data analysis, including time alignment and windowing, computation of clinically relevant glycemic metrics, and cohort-level visualization.

Many downstream applications, particularly machine-learning workflows, require time-aligned signals.
DIAX therefore includes utilities for resampling and interpolation that help users generate task-appropriate representations while minimizing artifacts from frequency changes.
Alignment is fully customizable per column: for example, a replay-simulation task may require a fully interpolated, gap-free 5-minute grid, whereas a bolus-advisor model may prefer a 15-minute grid that preserves missing data as NaNs.

We provide a robust MATLAB class supporting standardized computation of glycemic metrics (e.g., time-in-ranges) and AGP-style plots for individuals and cohorts, with flexible time-based slicing across days, weeks, or user-defined windows and optional export of results.
It stores data in native time-table structures and integrates directly with community tools such as AGATA \citep{cappon2024agata}, allowing users to apply external analysis workflows without additional preprocessing.

\begin{figure}[ht]
  \centering
  \begin{subfigure}[b]{0.48\textwidth}
    \centering
    \includegraphics[width=\textwidth]{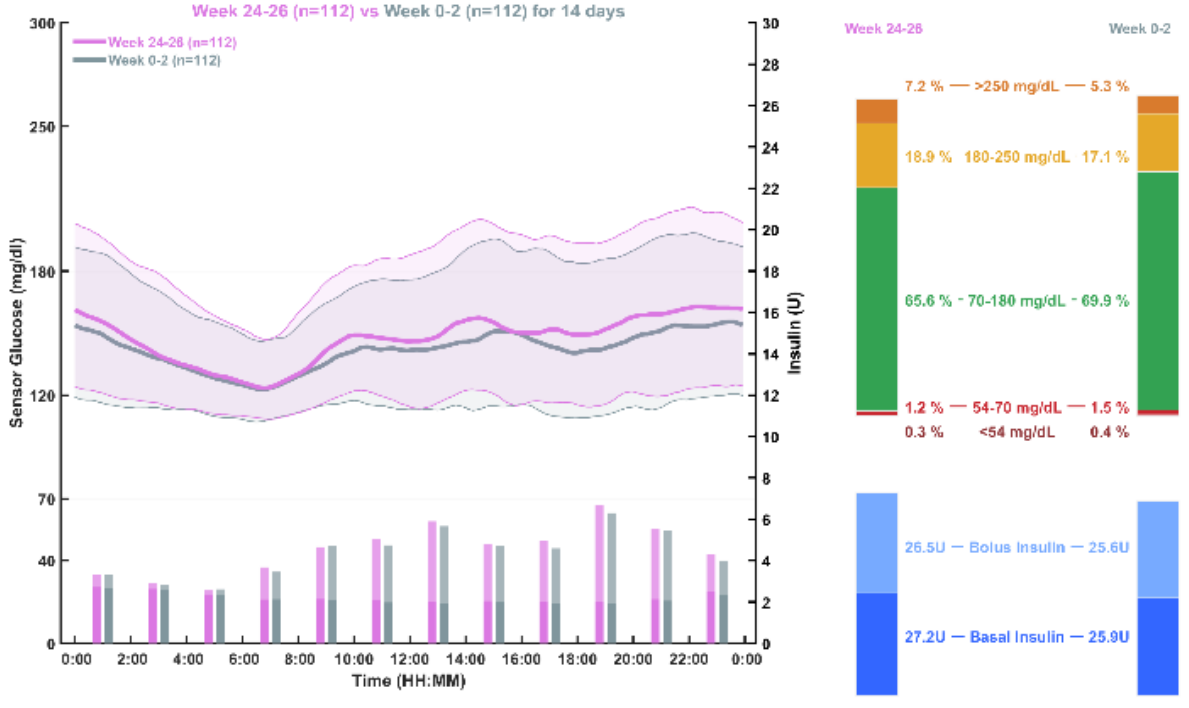}
  \end{subfigure}
  \hfill
  \begin{subfigure}[b]{0.48\textwidth}
    \centering
    \includegraphics[width=\textwidth]{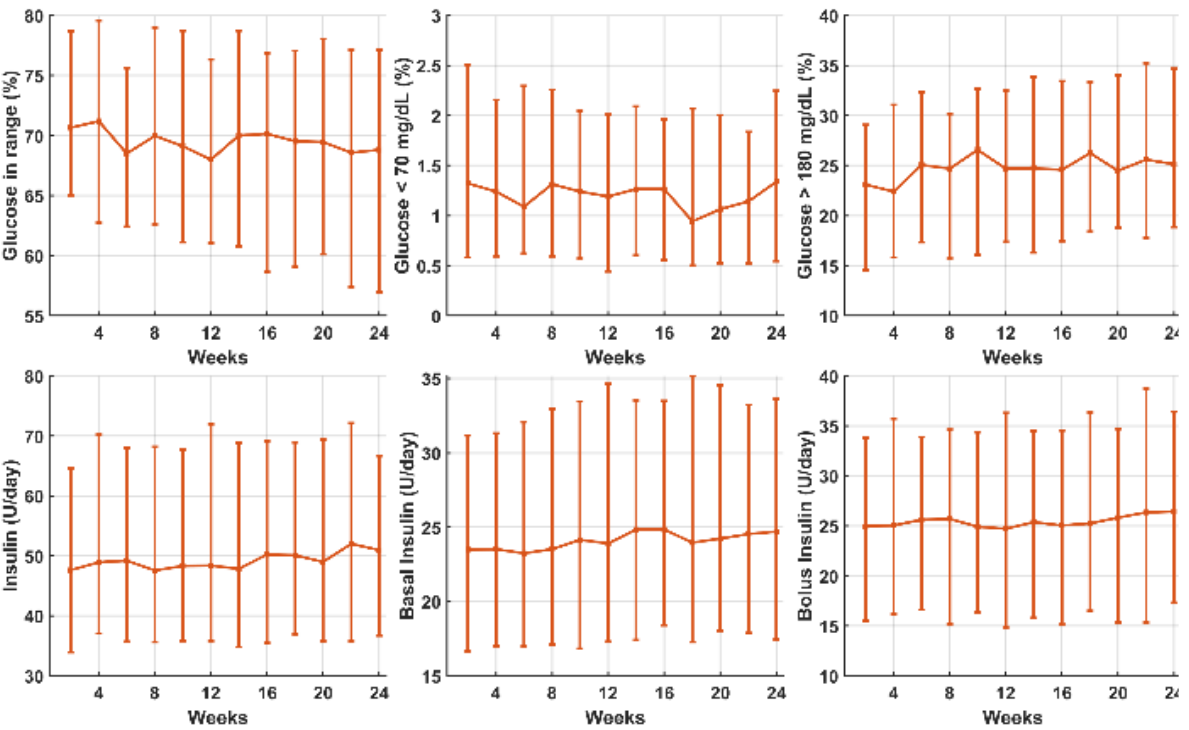}
  \end{subfigure}
  \caption{Example visualization of DCLP3 data.
  Left shows comparison of AGP for two weeks generated with the call \texttt{DIAX.plotCompare}.
  Right shows outcomes over time generated with the call \texttt{DIAX.plotOutcomes}.}
  \label{fig:visualization}
\end{figure}

Overall, the utilities are designed to give researchers convenient, ready-to-use tools for exploration, visualization, and metric computation.
Because all DIAX datasets share a unified schema, analysis code and tools developed for one study can be directly applied to others, enabling rapid inspection of new datasets and avoiding repeated dataset-specific reimplementation.

\section{Conclusion}

DIAX offers a flexible and transparent framework for representing diabetes-related time series data, emphasizing glucose and insulin dynamics.
It prioritizes interoperability, extensibility, and reproducibility, qualities that are valuable for machine learning applications, where large and diverse datasets are essential.
Additional utility functions using the DIAX format are also provided.
This can further reduce the burden required for basic data analysis or integration in downstream pipelines.

Beyond simplifying integration, a universal format creates a foundation for community-driven innovation.
By standardizing structure, researchers can develop and share data-driven methods, visualization libraries, and analysis tools that work seamlessly across datasets, reducing duplication of effort and accelerating progress.
This interoperability not only promotes transparency and reproducibility but also supports scalable research and open science, empowering the global diabetes technology community to collaborate more effectively.

While DIAX is broadly applicable and well-suited for general-purpose analysis, it is a research harmonization layer rather than a clinical data exchange format.
It is not a substitute for detailed clinical datasets; certain research questions may require returning to the original source data for full context.

By enabling community contributions, particularly from original dataset authors, the repository supports the inclusion of a wide range of datasets without the logistical burden of hosting raw data.
This approach fosters a richer and more diverse data ecosystem, empowering researchers to explore new questions, validate models across populations, and accelerate progress in diabetes technology and care.


\bibliographystyle{unsrtnat}
\bibliography{references}  

\end{document}